\documentclass{article} 
\usepackage{iclr2025_conference,times}
\usepackage{makecell}

\usepackage{amsmath,amsfonts,bm}

\def\eqref#1{equation~\ref{#1}}

\def\1{\bm{1}}

\DeclareMathAlphabet{\mathsfit}{\encodingdefault}{\sfdefault}{m}{sl}
\SetMathAlphabet{\mathsfit}{bold}{\encodingdefault}{\sfdefault}{bx}{n}

\usepackage{hyperref}
\usepackage{url}
\usepackage{booktabs}
\usepackage{graphicx}

\title{Identifying a Circuit for Verb Conjugation in GPT‑2}

\author{David Demitri Africa \thanks{ Also david.demitri.africa@gmail.com.} \\
Department of Computer Science\\
University of Cambridge\\
15 JJ Thomson Ave, Cambridge, United Kingdom CB3 0FD \\
\texttt{\{dda28\}@cam.ac.uk} \\ }

\iclrfinalcopy
\begin{document}

\maketitle

\begin{abstract}
I implement a procedure to isolate and interpret the sub‐network (or “circuit”) responsible for subject–verb agreement in GPT‑2 Small. In this study, the model is given prompts where the subject is either singular (e.g. “Alice”) or plural (e.g. “Alice and Bob”), and the task is to correctly predict the appropriate verb form (“walks” for singular subjects, “walk” for plural subjects). Using a series of techniques---including performance verification, automatic circuit discovery via direct path patching, and direct logit attribution---I isolate a candidate circuit that contributes significantly to the model’s correct verb conjugation. The results suggest that only a small fraction of the network’s component–token pairs is needed to achieve near‐model performance on the base task, but substantially more for more complex settings.

\end{abstract}
\textbf{Word Count:} 3999 \\
\textbf{Github:} https://github.com/DavidDemitriAfrica/verb-conjugation-mechinterp/
\section{Introduction}

Subject–verb agreement (SVA) is a fundamental aspect of English syntax. For example, given the inputs:
\begin{itemize}
    \item \textbf{Singular:} “Alice \_\_\_” should be completed with “walks”
    \item \textbf{Plural:} “Alice and Bob \_\_\_” should be completed with “walk”
\end{itemize}
GPT‑2 Small \citep{radford2019language} exhibits strong performance in this aspect, which is well known \citep{warstadt2020blimp, yu-etal-2020-word}. In this study, I conceptualize the task as predicting the correct verb conjugation based on the number of subjects provided in a prompt. My aim is to decompose the underlying computations into a circuit that maps from the subject token(s) to the correct verb form. I operationalize this by measuring the difference between the logit of the correct conjugation (e.g. “walks” or “walk”) and that of the incorrect form. Further, I iteratively build a circuit by composing more difficult settings on top of the base setting to understand how they are needed and when.

\section{Background}

\paragraph{Mechanistic Interpretability}
Mechanistic interpretability, often referred to as "mechinterp," has become an influential area within explainable artificial intelligence (XAI). It focuses explicitly on uncovering the internal computational processes of neural networks by decomposing them into interpretable sub-components such as neurons, attention heads, or circuits \citep{olah2020zoom, elhage2021mathematical}. Unlike other interpretability approaches that offer post-hoc explanations or approximate interpretations, mechanistic interpretability attempts to map explicit computational paths from model inputs to outputs \citep{casper2023mechanistic}. This technique has improved our understanding of transformer models significantly in terms of transparency, debugging, and trustworthiness \citep{meng2022locating, geiger2023causal}.

\paragraph{Circuits in Transformer Models}
The concept of a "circuit" has emerged prominently in the mechanistic interpretability literature, describing minimal subnetworks responsible for specific behaviors or functions within large neural networks \citep{wang2022interpretability, cammarata2020thread, olsson2022context}. Recent studies have revealed that complex behaviors exhibited by transformer-based models, such as semantic interpretation or syntactic processing, can often be traced back to surprisingly small groups of neurons or attention heads that significantly contribute to the model's output \citep{elhage2021mathematical, geva2022transformer}. Identifying such circuits not only clarifies how neural networks represent linguistic structure but also offers practical ways to intervene or refine model behavior in a sensible fashion \citep{meng2022locating}.

\paragraph{The Interpretability in the Wild Study}
One influential example of circuit discovery in transformers is the Indirect Object Identification (IOI) interpretability study conducted by \citet{wang2022interpretability}. This foundational work systematically demonstrated that transformer models rely on distinct internal subnetworks—or circuits—to accurately resolve indirect-object references in textual inputs. By employing path-patching and attribution methods, \citet{wang2022interpretability} identified specific attention heads responsible for processing and propagating linguistic context crucial to IOI tasks. Their techniques serve as the methodological foundation for the study.

\paragraph{GPT-2}
GPT-2, is a widely studied transformer-based language model introduced by \citet{radford2019language}, and has been probed for circuits in a variety of studies \cite{wang2022interpretability, hanna2023does, garcia2024does, yao2024knowledge, mathwin2023identifying}. I focus on GPT-2 Small, a decoder-only transformer with 12 layers and 12 heads for each attention layer.

\paragraph{Subject–Verb Agreement}
Subject–verb agreement is a fundamental area of linguistic inquiry, extensively studied in both theoretical linguistics and psycholinguistics due to its centrality in syntactic structure \citep{bock1991broken, eberhard2005making, franck2010object}. Agreement requires verbs to match their subjects in grammatical number and person, which means models have to understand the underlying syntactic relationships within sentences \citep{corbett2006agreement, franck2010object}. Historically, computational linguistic models have historically struggled with subject–verb agreement, making it a crucial benchmark for assessing syntactic capability in language models \citep{linzen2016assessing, goldberg2019assessing}. The relatively surprising performance of large language models is worth studying, both to understand the reasons for their performance and to construct more explainable computational linguistic models.

Briefly, I motivate bringing these components together. Subject–verb agreement provides a valuable domain for mechanistic interpretability because it requires the neural network to demonstrate abstract linguistic capabilities like number agreement, syntactic structure recognition, and morphological generalization—all fundamental components of natural language understanding. Unlike the IOI task studied by \citet{wang2022interpretability}, SVA explicitly tests a model's capacity to represent abstract grammatical categories (e.g., singular vs. plural) rather than just semantic or lexical relationships. SVA tasks thus allow us to probe how transformers handle syntactic generalization across diverse linguistic contexts systematically. Furthermore, examining SVA can reveal whether transformers rely on diffuse, redundant structures or compact, easily interpretable subnetworks to encode syntactic relations, helping us understand linguistic generalization within transformer architectures.

\section{Methodology}

\subsection{Verifying GPT‑2 on the Verb Conjugation Task}
The prerequisite to find a circuit in GPT-2 is to assess if the overall model can do the task correctly. If the model cannot do the task, then there is no point to look for a circuit that doesn't exist. While GPT-2 has been extensively studied in both a computational and linguistic sense, to my knowledge there has been no attempt to decompose GPT-2's performance in subject-verb agreement in this in-depth a fashion.

\textbf{Task:} I assess GPT‑2’s ability to correctly perform subject–verb agreement in controlled contexts. Each prompt is constructed from a combination of factors—including subject type (name vs. pronoun), subject plurality (singular vs. plural), sentence prefix (to add complexity, and distinguish time-specific or neutral), verb type (regular vs. irregular), tense (present vs. past), and negation—I provide a “prompt” ending in a context where the next-token should be the conjugated verb. For example, a prompt might be:
\[
\texttt{"Yesterday, Alice did not \_\_\_"}
\]
where the model is expected to predict the correct base form of the verb (e.g. “walk” in the case of a regular verb).

\textbf{Metric:} For each prompt, I compute:
\[
\text{Logit Difference} = \text{Logit}[\text{correct verb}] - \text{Logit}[\text{incorrect verb}],
\]
where a positive logit difference indicates that GPT‑2 assigns a higher probability to the correct conjugation. I aggregate this difference over 100 samples for each unique permutation of six controlled conditions (resulting in 6,400 examples overall). Classification metrics such as accuracy, precision, recall, and F1 across these conditions are computed as a natural extension of the logit difference metric; the answer with the higher logit is the prediction of the model.

\subsection{Controlled Permutations and Their Implications}
By systematically varying six conditions—\textit{is\_plural}, \textit{is\_negated}, \textit{has\_prefix}, \textit{is\_pronoun}, \textit{tense}, and \textit{use\_irregular}—I create a comprehensive grid of scenarios. This design allows me to assess the following settings and their implications on GPT-2's language reasoning capabilities upon failure.

\begin{table}[h]
\centering
\renewcommand{\arraystretch}{1.3} 
\begin{tabular}{|p{2.5cm}|p{6cm}|p{5.5cm}|}
\hline
\textbf{Setting} & \textbf{Example Prompt and Verb Forms} & \textbf{Failure Implication} \\ \hline
BASE & 
\texttt{Alice } \quad (correct: \texttt{walks} / incorrect: \texttt{walk}) & 
GPT-2 fails to reliably predict standard subject–verb agreement. \\ \hline
is\_plural & 
\texttt{Alice and Bob } \quad (correct: \texttt{walk} / incorrect: \texttt{walks}) & 
GPT-2 fails to account for plural subjects. Does not understand proper agreement and abstraction over number. \\ \hline
is\_negated & 
\texttt{Alice does not } \quad (correct: \texttt{walk} / incorrect: \texttt{walks}) & 
GPT-2 fails to interpret negation cues are misinterpreted, or auxiliary-driven base forms are overlooked. \\ \hline
has\_prefix & 
\texttt{Surprisingly, Alice } \quad (correct: \texttt{walks} / incorrect: \texttt{walked}) & 
GPT-2 fails to generalize to slightly more complex sentence structure. \\ \hline
is\_pronoun & 
\texttt{She } \quad (correct: \texttt{walks} / incorrect: \texttt{walk}) & 
GPT-2 fails to understand pronoun substitution, reflecting shallow handling of subject type. \\ \hline
tense & 
\texttt{Alice } \quad (in past: correct: \texttt{walked} / incorrect: \texttt{walk}) & 
GPT-2 fails to adapt to temporal contexts. \\ \hline
use\_irregular & 
\texttt{Alice } \quad (irregular verb example: correct: \texttt{eats} / incorrect: \texttt{eat}) & 
GPT-2 generically processes lexical forms and does not generalize to irregular verb forms. \\ \hline
\end{tabular}
\caption{Controlled permutations with representative examples and failure implications.}
\label{tab:controlled_settings_actual}
\end{table}

\subsection{Path Patching} 
Path patching is a causal intervention technique in which the activation of a particular network component (e.g. an attention head) is replaced with the activation computed from a counterfactual input. This substitution isolates the contribution of that component by measuring the change in the model's output \citep{wang2022interpretability, olah2020zoom, goldowskydill2023localizingmodelbehaviorpath}.

For each attention head in GPT-2, I compute the \textit{head effect} defined as the difference between the baseline logit difference (computed from the original prompt) and the logit difference after \textit{patching} the head's activation with that from a counterfactual input. The counterfactual is generated by flipping some condition in the setting (such as the subject's plurality) thereby swapping the correct and incorrect verb forms. I use forward hooks on the attention modules to capture and replace activations.

To replace the activation, there are three well-known ablation strategies:
\begin{itemize}
    \item \textbf{Zero Ablation:} The head's output is replaced with a tensor of zeros. This method is straightforward, but also the most naive as it may not preserve a bias term passed down by default in the network.
    \item \textbf{Mean Ablation:} The head's output is replaced with its mean activation (computed over a sample of prompts). This method from \citet{meng2022locating} is intended to neutralize the head's dynamic contribution while preserving its baseline signal.
    \item \textbf{Resample Ablation:} The head's output is replaced by a randomly selected activation from a pool of pre-collected samples. This is suggested by \citet{geiger2023causal} as zero and mean activations take the model too far away from actually possible activation distributions, and is used in a wide amount of interpretability projects \citep{hanna2023does, wang2022interpretability, conmy2023automated} as pointed out by \citet{conmy2023automated}.
\end{itemize}

As resample ablation is the most realistic and well-supported, I use it to perform a knockout experiment whereby all heads not in a candidate circuit are ablated. The performance of the circuit is then evaluated by comparing the average logit difference (i.e. the model’s confidence in the correct conjugation) of the circuit against that of the full model.

\subsection{Greedy Iterative Circuit Search} 
After computing per-head effects via path patching, I rank attention heads by their absolute contribution. I greedily sample node-by-node starting from the highest ranked attention heads, and take in heads which improve the accuracy of the preexisting circuit. I stop when circuit performance no longer meaningfully improves (\(F_{\mathrm{circuit}}\)) within a specified tolerance relative to the full model’s performance (\(F_{\mathrm{full}}\)). This method approximates a circuit that is sufficient for SVA, in line with the principles outlined by \citet{wang2022interpretability}. I note, however, that this greedy search may omit components such as the \textit{Negative Name Mover heads}, which reduce the confidence of the model to hedge against uncertainty.

To find the base minimum circuit, I perform this on the base setting with resample ablation to identify a circuit that yields performance close to the full model, and excluding unhelpful heads.

\subsection{Iterative Circuit Expansion}
Once the base circuit is established, I extend the analysis to the more challenging linguistic conditions. For each additional setting (e.g., negation, pronoun substitution, presence of a prefix, irregular verbs, or the full unfiltered dataset), I:
\begin{enumerate}
    \item Evaluate the full model’s performance on the setting.
    \item Evaluate the performance of the base circuit (obtained on the base dataset) on the same setting.
    \item Iteratively search for and add extra heads from the candidate pool that significantly reduce the performance gap relative to the full model.
\end{enumerate}

\section{Results and Discussion}

I find that GPT-2 is successful in identifying the correct verb conjugation to at least a reasonable degree of accuracy in every setting and permutation thereof. I present the results plainly below. The full detailed table with all permutations can be found in the appendix as Table~\ref{tab:appendix_full}. 

\begin{table}[htbp]
\centering
\begin{tabular}{|l|r|r|}
\toprule
\textbf{Setting} & Accuracy & F1 Score \\
\midrule
BASE  & 0.63 & 0.77 \\
is\_plural & 0.73 & 0.77 \\
is\_negated  & 0.90  & 0.95 \\
has\_prefix  & 0.74 & 0.85 \\
is\_pronoun  & 0.69 & 0.82 \\
tense\_past & 0.62 & 0.77 \\
use\_irregular & 0.62 & 0.77 \\
ALL & 1.00 & 1.00 \\
\bottomrule
\end{tabular}
\caption{Model Performance under Incremental Conditions (Summary)}
\label{tab:main_summary}
\end{table}

While not the focus of the study, it is interesting that GPT-2 has the highest performance in the most complex setting, which \textit{a priori} should be most difficult to model. 

Now I examine the average effects of each attention head on the task.

\begin{figure}
    \centering
    \includegraphics[width=0.66\linewidth]{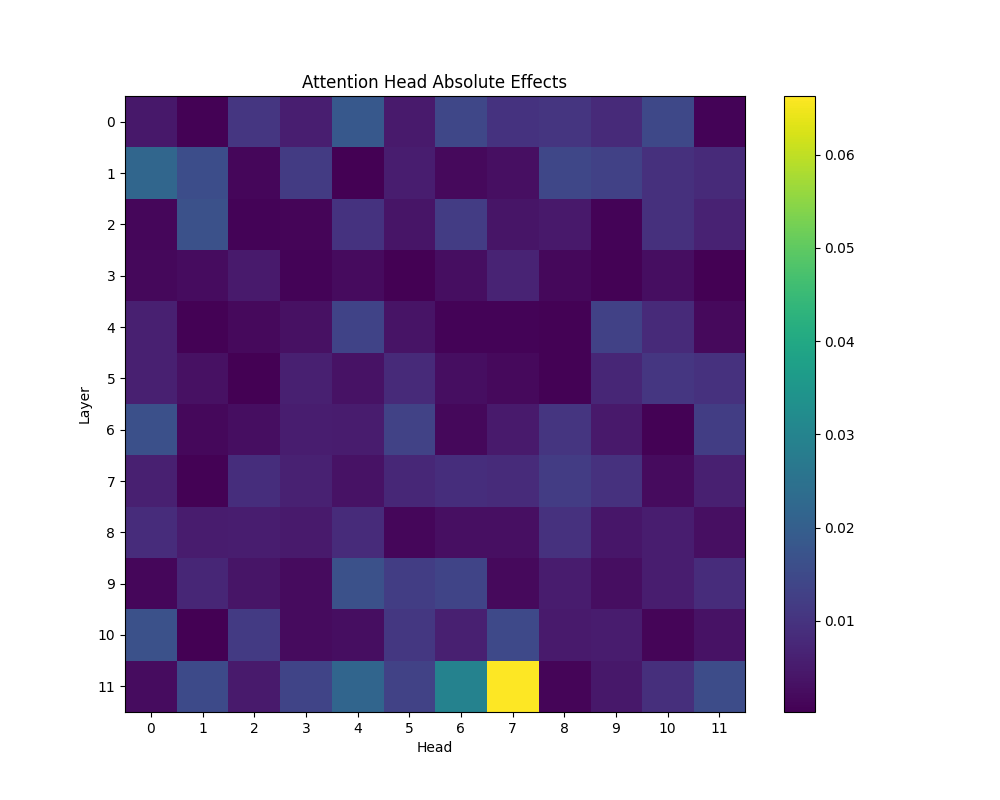}
    \caption{Heatmap of average attention head effects on the base subject–verb agreement task. Positive values (yellow) indicate heads crucial to correct verb conjugation, negative values (purple) suggest inhibitory roles, and values near zero (green) represent heads with minimal direct impact.}
    \label{fig:attn_head_effects}
\end{figure}

Figure~\ref{fig:attn_head_effects} displays the absolute effects of each attention head in GPT-2 Small (12 layers, 12 heads per layer), computed via activation patching. Head 7 in layer 11 (bright yellow) exerts the most substantial influence on the model’s predictions. The second and fourth most influential heads are also in layer 11, which may suggest that the processing of contextual syntactic information is primarily done in the late layers of the model. The majority of other heads exhibit relatively minor effects (dark purple to blue shades), consistent with previous studies indicating that linguistic behaviors often rely on sparse circuits within large transformer models \citep{wang2022interpretability, conmy2023automated, geiger2023causal}.

\subsection{Base Circuit}

I first identify a base circuit capable of performing the simplest form of the subject–verb agreement task (the "base" setting), where the model predicts the correct verb conjugation given a singular named subject. This minimal circuit consists of only 12 attention heads spanning 7 layers, a highly compact representation requiring only a twelfth of the full model size. This basic circuit achieves an accuracy of 0.65, which is close to the full model’s accuracy of 0.70.

\subsection{Expanded Circuits}

Subsequently, I iteratively expand the base circuit to handle more linguistically complex scenarios. I track which and how many additional attention heads are needed to match or closely approach the full model's performance. Table~\ref{tab:circuit_expansion} summarizes the expansions and associated accuracy results:

\begin{table}[h]
\centering
\begin{tabular}{lcc}
\toprule
Setting & Heads in Expanded Circuit & Circuit Accuracy / Full Model Accuracy \\
\midrule
Base & 12 & 0.65 / 0.70 \\
Plural & 23 & 0.53 / 0.88 \\
Negation & 29 & 0.53 / 0.91 \\
Prefix & 82 & 0.78 / 0.91 \\
Pronoun & 22 & 0.57 / 0.88 \\
Past Tense & 55 & 0.63 / 0.86 \\
Irregular & 97 & 0.82 / 0.93 \\
Complex & 125 & 0.80 / 1.00 \\
\bottomrule
\end{tabular}
\caption{Iterative circuit expansions and corresponding performance on various linguistic settings.}
\label{tab:circuit_expansion}
\end{table}

The iterative circuit expansions revealed clear patterns.

\paragraph{Poor circuit performance.} Three settings—plural, negation, and pronoun—stood out due to their notably poor circuit performance, despite modest expansions. For the plural setting, an addition of just 11 heads yielded minor improvement in accuracy (0.53) relative to the full model’s significantly higher accuracy (0.88). This suggests that plural agreement might be more diffusely encoded across multiple model components beyond just the attention heads (or even in the MLP layers \cite{yao2024knowledge}.

The negation setting similarly required a modest addition of 17 heads, yet the circuit accuracy (0.53) remained markedly below the full model’s accuracy (0.91). Negation is known to be broadly distributed across neurons from mechanistic interpretability \citep{meng2022locating, geiger2023causal} and difficult to model properly from computational linguistics literature \citep{truong-etal-2023-language, kassner-schutze-2020-negated}.

Introducing pronouns required an even smaller expansion of just 10 heads. Yet, the resultant circuit accuracy (0.57) was considerably lower than the full model’s accuracy (0.88). Again, pronoun processing in GPT-2 might use non-attention or distributed mechanisms in the same way for the plural setting.

In general, I would be skeptical about any of the conclusions drawn from poor performance. There are many reasons why a subgraph might do poorly, such as the greedy search not looking at all subset interactions, and missing out on non-linear interactions. This is one reason that would explain why simpler tasks paradoxically yield worse performance: they have fewer redundancies, making them vulnerable to small subgraph omissions during greedy circuit discovery. Future analysis involving more comprehensive optimization methods (such as ACDC) could confirm the reason for this odd behavior.

\paragraph{Comparable model accuracy.} The circuit achieves significant accuracy gains in the remaining settings: prefix, past tense, irregular, and complex. The most obvious commonality between them is that they require significantly more heads than the base circuit, which likely explains the increased performance. It may be that since these settings do not flip the singular/plural conjugation in the way that the plural and negation settings do, so the greedy search successfully found more useful heads.

Now, I suggest possible reasons why such large expansions were needed to reach an appreciable fraction of the full model accuracy. The prefix setting required a large expansion (70 additional heads) to achieve a circuit accuracy of 0.78 (close to the full model’s 0.91). The large number of heads needed suggests that GPT-2 relies heavily on widespread contextual information across numerous layers to correctly handle prefixes. Prefixes add semantic and syntactic complexity, requiring the model to integrate multiple context-dependent cues; hence, this complexity naturally leads to a larger subnetwork.

The past tense verbs required a moderate expansion of 43 heads to achieve an accuracy of 0.63 versus the full model’s accuracy of 0.86. This suggests past tense processing involves moderately sized, specialized subnetworks. Past tense agreement, which depends heavily on temporal cues, might be encoded within more localized subnetworks compared to prefix handling, reflecting temporal reasoning mechanisms that span multiple model layers but still remain somewhat localized.

The irregular verbs required a significant expansion significantly (adding 85 heads), achieving an impressive circuit accuracy of 0.82 compared to the full model’s 0.93. Irregular verbs rely on extensive morphological and lexical knowledge encoded throughout numerous attention heads. Irregular forms cannot rely solely on generalizable syntactic rules and instead require memorization or pattern recognition at the lexical level, and are not easy to abstract. Hence it makes sense that such a large circuit is required.

Finally, the most complex setting, integrating all previous settings, required the largest circuit expansion (125 heads) to achieve a circuit accuracy of 0.80, still noticeably below the full model’s perfect accuracy of 1.00. This setting suggests that GPT-2's full capacity and strong performance arise from coordinated interactions among numerous specialized components rather than isolated computations within minimal subnetworks. At this point, the circuit is comprised of the majority of the model, and the ablated heads mostly serve redundant or irrelevant purposes.

\begin{figure}
    \centering
    \includegraphics[width=1\linewidth]{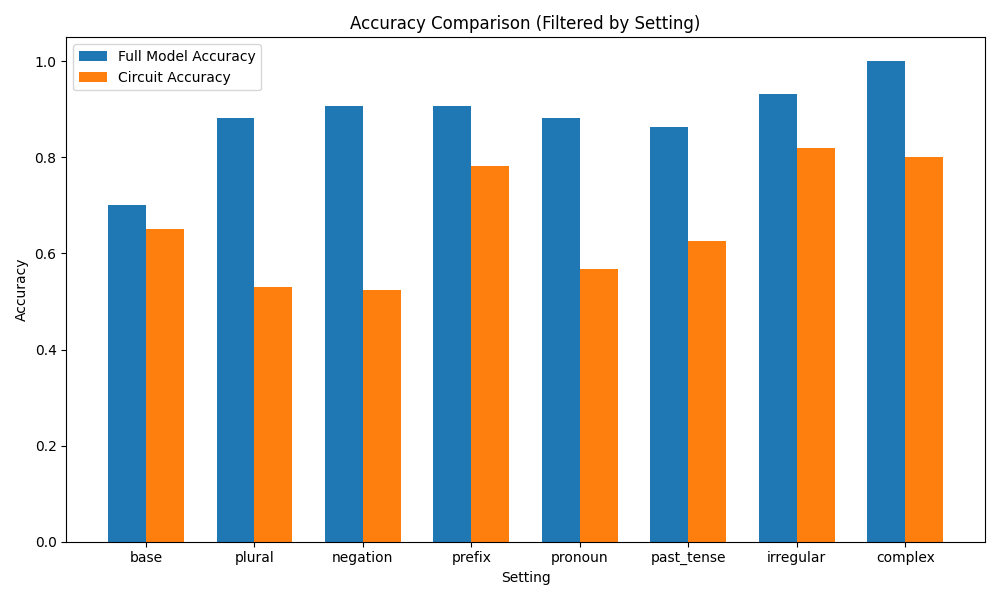}
    \caption{Accuracy comparison between the full GPT-2 model and the iteratively expanded circuits across different linguistic settings. The growth in accuracy seems to correlate directly with the size and complexity of the circuits, especially in more challenging linguistic contexts.}
    \label{fig:accuracy_comparison}
\end{figure}

The accuracy comparison in Figure~\ref{fig:accuracy_comparison} illustrates clearly how increased circuit size correlates strongly with improved accuracy, particularly as linguistic complexity increases. The unexpectedly high accuracy in the complex scenario (all settings combined) compared to simpler individual linguistic variations warrants further explanation. One plausible explanation is that, in scenarios that combine multiple linguistic complexities (plural, negation, irregularity, and temporal context), GPT-2 may leverage additional redundant circuits or internal mechanisms, becoming robust through broader coverage by a larger number of attention heads. In contrast, simpler scenarios (such as plural-only or negation-only) might unexpectedly stress more specialized circuits with fewer redundancies, causing failures more visibly when ablating parts of the network. This paradoxically leads to lower observed performance when only subsets of complexity are tested. Another factor could be dataset artifacts: the diversity of linguistic cues in combined scenarios might provide stronger, clearer signals to the model than individual simpler cues, inadvertently simplifying the internal computations required by the network.

\subsection{Shared Heads}

I also look at which settings shared heads. For instance, several key heads from the base circuit—particularly layer 11 heads 6, 4, and 7, and layer 0 heads 4 and 8— are consistently involved across all settings. When ablated, model performance drops drastically, which suggests these heads might be responsible for core syntactic and morphological processing tasks fundamental to verb agreement, regardless of linguistic variation. 

On the other hand, the significant overlap between the irregular and complex scenarios (with substantial sharing of heads from layers 7 and 9) suggests that handling irregular verb forms and complex combinations of linguistic features requires similar underlying computational mechanisms. This makes intuitive sense since irregular verb forms often involve lexical memorization or nuanced morphological rules, which would dominate in the final complex setting as irregular forms directly determine the final verb conjugation.

In general, these shared mechanisms are both logical and intriguing, and it would be interesting to look more into the structural mechanisms attended to by these heads.

\subsection{Head Functions}

I focus here on the base circuit of twelve heads that emerged from our greedy search on the simplest SVA tasks, since these heads reappear consistently in expansions and tend to be the largest contributors to performance. I conduct a limited investigation on the sentences "Alice walks", "*Alice walk", "Alice and Bob walk", and "*Alice and Bob walks", looking at a heatmap of the key value of the verb and the query value of each other token. Due to how time consuming the manual investigation of each head is, it is difficult to do this for other settings and for the expanded circuit. This is not a conclusive analysis, but is intended to provide an insight for what the partial functions of certain GPT-2 attention heads are in the context of subject verb agreement. 

To do this, I look at the attention heatmap of each head where each cell $(i,j)$ shows how strongly the token at position $i$ (query) attends to the token at position $j$ (key).

Very loosely, I identify four overarching behaviors in Table \ref{tab:head_categories}.

\begin{table}[h]
\centering
\resizebox{\textwidth}{!}{%
\begin{tabular}{p{4cm} p{7cm} p{5cm}}
\toprule
\textbf{Category} & \textbf{Description (Attention Patterns)} & \textbf{Heads in Base Circuit} \\
\midrule
\textit{Primary Subject-Anchor} &
Queries from verb positions strongly attend to keys at subject positions (e.g., "Alice"), forming a direct dependency between subject and verb tokens. High sensitivity to plurality mismatch, reflected in significant attention changes for singular vs. plural prompts. &
(2,1), (2,6), (9,4), (10,0), (11,4), (11,6), (11,7) \\[8pt]

\textit{Diffuse Subject-Scanner} &
Queries from verb positions distribute attention broadly over all prior positions (subjects and occasionally the verb), with only minor shifts in attention when subjects differ in plurality. &
(0,8), (1,0) \\[8pt]

\textit{Conjunction-Tracking} &
Queries from verb or plural subject positions strongly attend to keys at the conjunction token ("and"), indicating a clear role in handling compound subjects. Especially pronounced in plural contexts. &
(1,1) \\[8pt]

\textit{Invariant} &
Self-attention dominant pattern, where queries from tokens primarily attend to their own positions (diagonal), or uniformly distribute attention independent of subject–verb agreement conditions. &
(0,4), (6,0) \\
\bottomrule
\end{tabular}}
\caption{Categorization of heads in the base circuit based on attention patterns observed between query and key positions.}
\label{tab:head_categories}
\end{table}

\begin{figure}[htbp]
  \centering
  \begin{minipage}[b]{0.45\textwidth}
    \centering
    \includegraphics[width=\textwidth]{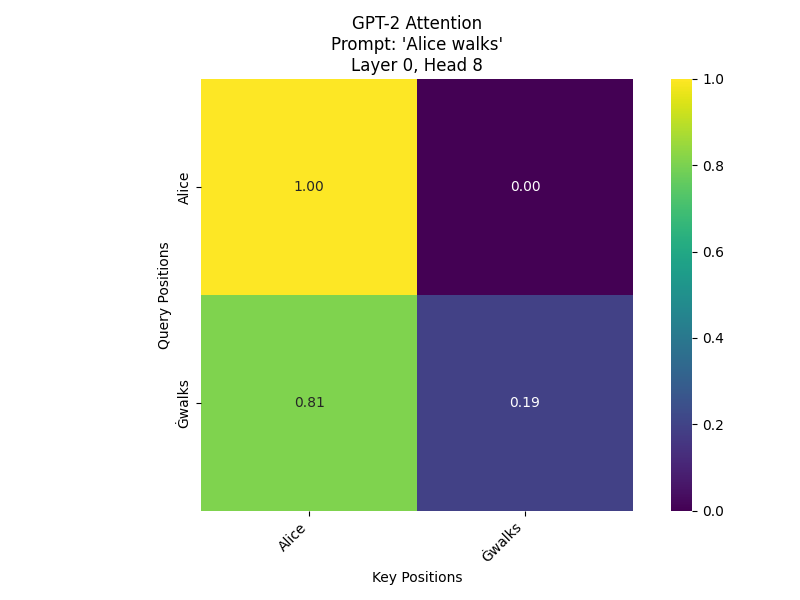}
    
    {\small (a) Alice walks layer0 head8}
  \end{minipage}
  \hfill
  \begin{minipage}[b]{0.45\textwidth}
    \centering
    \includegraphics[width=\textwidth]{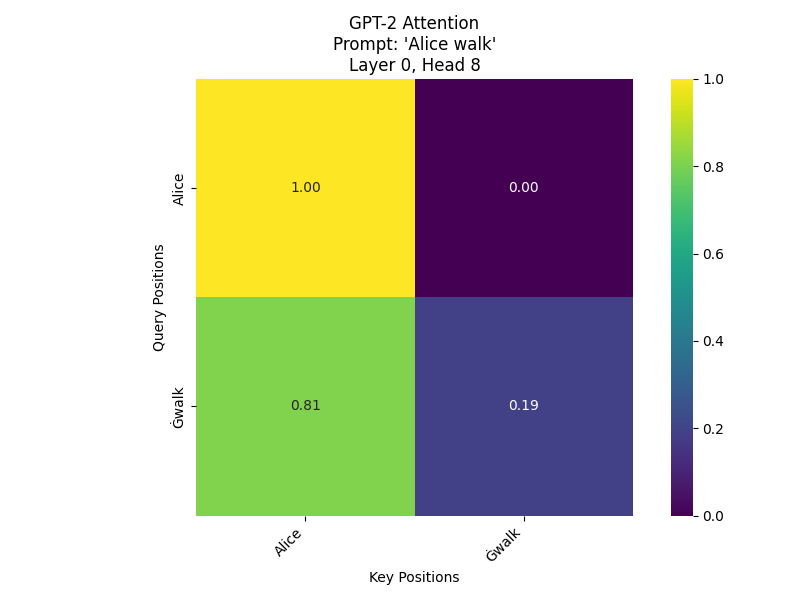}
    
    {\small (b) *Alice walk layer0 head8}
  \end{minipage}
  
  \vspace{1em} 
  
  \begin{minipage}[b]{0.45\textwidth}
    \centering
    \includegraphics[width=\textwidth]{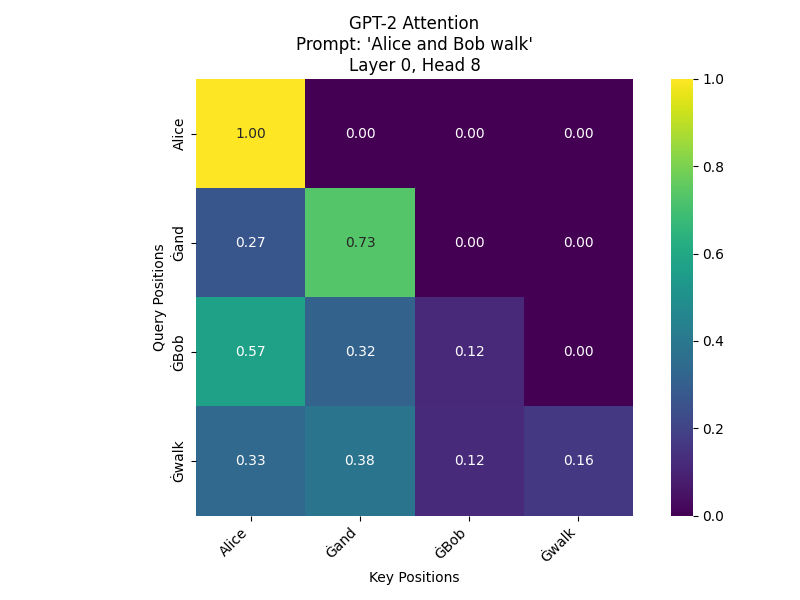}
    
    {\small (c) Alice and Bob walk layer0 head8}
  \end{minipage}
  \hfill
  \begin{minipage}[b]{0.45\textwidth}
    \centering
    \includegraphics[width=\textwidth]{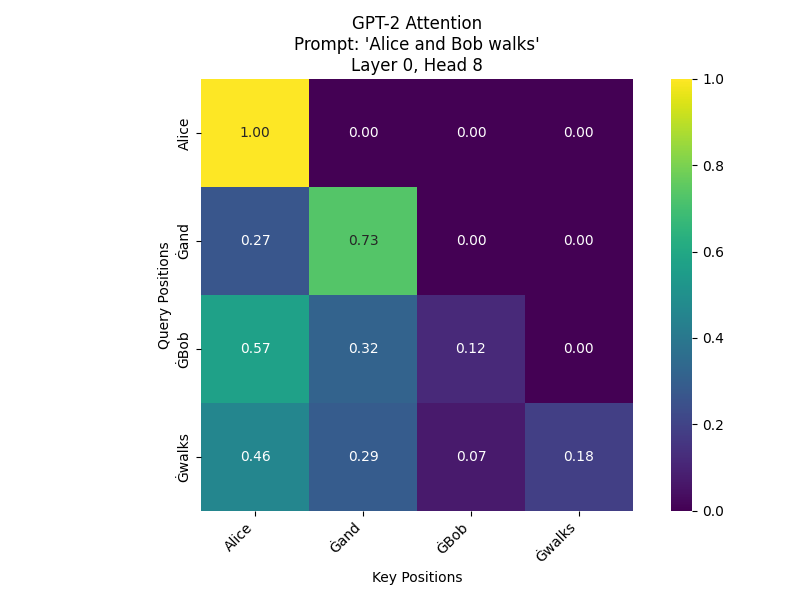}
    
    {\small (d) *Alice and Bob walks layer0 head8}
  \end{minipage}
  
  \caption{Attention heatmaps illustrating a representative diffuse subject-scanner (layer 0, head 8). Each heatmap cell indicates attention strength from query tokens (vertical axis) to key tokens (horizontal axis). This head consistently distributes attention broadly across previous tokens, exhibiting some sensitivity to grammatical correctness (e.g., singular vs. plural agreement). (a) and (b) depict singular subjects with correct ("Alice walks") and incorrect ("Alice walk") conjugation, respectively, while (c) and (d) show plural subjects with correct ("Alice and Bob walk") and incorrect ("Alice and Bob walks") conjugation, respectively. Attention patterns remain broadly distributed in all scenarios.}
  \label{fig:2x2grid}
\end{figure}

There is some weak linguistic justification for these categorizations: Primary Subject-Anchors might reflect direct dependency recognition (akin to syntactic parsing heads), while Diffuse Subject-Scanners might correspond to general contextual aggregation functions. We visualize an example of one such head in Figure \ref{fig:2x2grid}.

In general, it seems that many heads consistently exhibit a bias toward the initial token ("Alice"), regardless of grammatical structure. This phenomenon is particularly prevalent in higher layers (e.g., Layer 9, Layer 10, Layer 11).

Verb form also seems to minimally influences attention distribution, which may be a result of how the circuit is selected, optimizing for performance on attending to the verb form rather than from the verb form (as the verb is obscured in the SVA task).

Early layers (Layer 0-2) showed more diverse attention distributions, whereas layers (Layers 6–11) increasingly converge on strongly attending to the first token, suggesting these layers abstract toward a more general positional or topic-focused attention. It is entirely possible that the heads are just attending to the first token or possibly a proper noun, rather than having anything to do with SVA.

\section{Conclusion}

In this study, I implemented a systematic approach to isolate and interpret a candidate circuit responsible for subject–verb agreement in GPT-2 Small. My methodological choices closely followed the original interpretability-in-the-wild paper intentionally to maintain consistency and comparability. Specifically, resample ablation was used because, as Wang et al. argued, it preserves realistic activation distributions, avoiding artificially induced activations (zero-ablation) or overly simplistic assumptions (mean-ablation). Likewise, the artificial dataset structure was chosen deliberately to systematically control linguistic variables, enabling precise causal attributions of attention heads to linguistic features. This systematic control allows clearer interpretation and attribution of circuit components than would be possible with naturalistic, uncontrolled data. Finally, the greedy iterative circuit search was selected primarily because it provides computational tractability given the strict time constraints (80-hour scope), even though this approach might overlook nonlinear interactions.

I identified a minimal subnetwork capable of accurately performing certain verb conjugation tasks. The analysis revealed that GPT-2 relies predominantly on a small subset of attention heads, particularly concentrated in later layers, for basic syntactic processing. However, as linguistic complexity increased, the size and scope of circuits expanded significantly, indicating more distributed processing for nuanced linguistic features such as negation, irregular verbs, and context-rich prefixes. The unexpected superior performance of GPT-2 on the combined complexity scenario may indicate that the model benefits from richer, redundant linguistic contexts. While individual complexities (like plurality alone) may isolate very specific sub-circuits, combined complexities could activate multiple redundant pathways, providing robustness against ablations.

\subsection{Limitations}
The primary methodological limitations in this analysis arise from several sources.

\textbf{Greedy Search Strategy}: The iterative, greedy selection of attention heads does not guarantee an optimal circuit, potentially missing important combinatorial interactions or redundant mechanisms. While it would be pleasant to speculate on the generality of the SVA heads here, there is no guarantee that all the heads are strictly necessary for the circuit to function nor if all heads which perform this function in the model were found. The search strategy, while practical and computationally tractable, overlooks subnetworks that only emerge in specific combinational contexts.

\textbf{Attention-Only Analysis}: The analysis focused exclusively on attention heads, omitting multi-layer perceptron (MLP) components entirely. Previous literature (e.g., \citealt{geva2022transformer, yao2024knowledge}) has demonstrated that MLP layers play crucial roles in syntactic and semantic processing, suggesting that the circuits identified here represent only partial explanations of GPT-2’s verb conjugation capabilities.

\textbf{Path Patching Limitations}: Although resample ablation is considered the most realistic intervention, it may still introduce unrealistic activation patterns. Ablations might cause cascading effects that exaggerate the importance of certain heads or underrepresent others, resulting in skewed interpretations.

\textbf{Limited Linguistic Coverage}: Despite extensive controlled variations, the linguistic settings explored remain bounded and artificial. Real-world syntactic complexity often combines multiple factors simultaneously, and model behavior under such naturally occurring conditions remains unexplored within this analysis. While it is heartening that the circuit performed well in the most realistic and complex setting, it is confusing and unexplained why it performed substantially worse in settings where only one complexifying setting was imposed.

\textbf{Manual Interpretation Bias}: The qualitative, manual categorization of head functions (e.g., subject-anchoring, conjunction-tracking) introduces subjectivity, which limits the generalizability and reproducibility of interpretations. This step of mechanistic interpretability is largely unstudied, as while circuit discovery has been automated, the actual interpretation of circuit components is still mostly subjective.

\subsection{Future Work}
Several directions emerge naturally from this study:

\textbf{MLP Integration}: Expanding the analysis to include MLP layers would provide a more complete understanding of the network's computational mechanics, potentially revealing additional mechanisms relevant to verb conjugation.

\textbf{Global Optimization Methods}: Employing optimization methods such as genetic algorithms or integer linear programming to identify globally optimal or near-optimal circuits could improve circuit identification beyond greedy methods. At minimum, it would make sense to implement ACDC on this dataset.

\textbf{Cross-Model Comparisons}: Replicating this circuit analysis across different transformer models or architectures would help us understand syntactic functions generalize across architectures and model scales.

In conclusion, this study provides both foundational insights and methodological tools to better understand transformer-based language models' internal syntactic reasoning mechanisms. This serves as a foundation for future interpretability research and how syntactic ability develops in models.

\bibliography{iclr2025_conference}
\bibliographystyle{iclr2025_conference}

\appendix
\section{Appendix}

\begin{table}
\subsection{Supplementary Figures}
\resizebox{\textwidth}{!}{%
\begin{tabular}{llllllrrrrrrr}
\toprule
Number & Negation & Prefix & Subject & Tense & Verb Type & Accuracy & Precision & Recall & F1 Score & Mean Logit Diff & Std Logit Diff & Count \\
\midrule
Singular & Affirmative & Without & Name & Past & Regular & 0.62 & 1.00 & 0.62 & 0.77 & 0.44 & 1.47 & 100 \\
Singular & Affirmative & Without & Name & Past & Irregular & 0.70 & 1.00 & 0.70 & 0.82 & 0.79 & 1.12 & 100 \\
Singular & Affirmative & Without & Name & Present & Regular & 0.63 & 1.00 & 0.63 & 0.77 & 0.46 & 1.73 & 100 \\
Singular & Affirmative & Without & Name & Present & Irregular & 0.62 & 1.00 & 0.62 & 0.77 & 0.11 & 0.38 & 100 \\
Singular & Affirmative & Without & Pronoun & Past & Regular & 0.81 & 1.00 & 0.81 & 0.90 & 1.40 & 1.66 & 100 \\
Singular & Affirmative & Without & Pronoun & Past & Irregular & 0.67 & 1.00 & 0.67 & 0.80 & 1.51 & 1.51 & 100 \\
Singular & Affirmative & Without & Pronoun & Present & Regular & 0.62 & 1.00 & 0.62 & 0.77 & 0.75 & 1.94 & 100 \\
Singular & Affirmative & Without & Pronoun & Present & Irregular & 1.00 & 1.00 & 1.00 & 1.00 & 0.74 & 0.45 & 100 \\
Singular & Affirmative & With & Name & Past & Regular & 0.94 & 1.00 & 0.94 & 0.97 & 3.44 & 2.11 & 100 \\
Singular & Affirmative & With & Name & Past & Irregular & 1.00 & 1.00 & 1.00 & 1.00 & 4.12 & 2.00 & 100 \\
Singular & Affirmative & With & Name & Present & Regular & 0.69 & 1.00 & 0.69 & 0.82 & 1.31 & 2.09 & 100 \\
Singular & Affirmative & With & Name & Present & Irregular & 1.00 & 1.00 & 1.00 & 1.00 & 1.67 & 0.76 & 100 \\
Singular & Affirmative & With & Pronoun & Past & Regular & 0.95 & 1.00 & 0.95 & 0.97 & 3.26 & 2.05 & 100 \\
Singular & Affirmative & With & Pronoun & Past & Irregular & 1.00 & 1.00 & 1.00 & 1.00 & 4.60 & 1.99 & 100 \\
Singular & Affirmative & With & Pronoun & Present & Regular & 0.85 & 1.00 & 0.85 & 0.92 & 2.05 & 2.17 & 100 \\
Singular & Affirmative & With & Pronoun & Present & Irregular & 1.00 & 1.00 & 1.00 & 1.00 & 2.39 & 0.60 & 100 \\
Singular & Negated & Without & Name & Past & Regular & 0.85 & 1.00 & 0.85 & 0.92 & 2.52 & 2.15 & 100 \\
Singular & Negated & Without & Name & Past & Irregular & 0.92 & 1.00 & 0.92 & 0.96 & 1.68 & 1.30 & 100 \\
Singular & Negated & Without & Name & Present & Regular & 0.90 & 1.00 & 0.90 & 0.95 & 2.94 & 2.14 & 100 \\
Singular & Negated & Without & Name & Present & Irregular & 1.00 & 1.00 & 1.00 & 1.00 & 3.52 & 0.29 & 100 \\
Singular & Negated & Without & Pronoun & Past & Regular & 0.90 & 1.00 & 0.90 & 0.95 & 2.53 & 2.10 & 100 \\
Singular & Negated & Without & Pronoun & Past & Irregular & 1.00 & 1.00 & 1.00 & 1.00 & 2.18 & 1.22 & 100 \\
Singular & Negated & Without & Pronoun & Present & Regular & 0.83 & 1.00 & 0.83 & 0.91 & 2.90 & 2.77 & 100 \\
Singular & Negated & Without & Pronoun & Present & Irregular & 1.00 & 1.00 & 1.00 & 1.00 & 4.01 & 0.19 & 100 \\
Singular & Negated & With & Name & Past & Regular & 0.79 & 1.00 & 0.79 & 0.88 & 1.83 & 2.17 & 100 \\
Singular & Negated & With & Name & Past & Irregular & 0.86 & 1.00 & 0.86 & 0.92 & 1.80 & 1.38 & 100 \\
Singular & Negated & With & Name & Present & Regular & 0.94 & 1.00 & 0.94 & 0.97 & 3.63 & 2.11 & 100 \\
Singular & Negated & With & Name & Present & Irregular & 1.00 & 1.00 & 1.00 & 1.00 & 4.02 & 0.22 & 100 \\
Singular & Negated & With & Pronoun & Past & Regular & 0.84 & 1.00 & 0.84 & 0.91 & 2.17 & 2.16 & 100 \\
Singular & Negated & With & Pronoun & Past & Irregular & 0.85 & 1.00 & 0.85 & 0.92 & 1.57 & 1.37 & 100 \\
Singular & Negated & With & Pronoun & Present & Regular & 0.89 & 1.00 & 0.89 & 0.94 & 3.06 & 2.30 & 100 \\
Singular & Negated & With & Pronoun & Present & Irregular & 1.00 & 1.00 & 1.00 & 1.00 & 3.82 & 0.24 & 100 \\
Plural & Affirmative & Without & Name & Past & Regular & 0.73 & 1.00 & 0.73 & 0.84 & 1.19 & 1.67 & 100 \\
Plural & Affirmative & Without & Name & Past & Irregular & 0.64 & 1.00 & 0.64 & 0.78 & 1.25 & 1.56 & 100 \\
Plural & Affirmative & Without & Name & Present & Regular & 0.73 & 1.00 & 0.73 & 0.84 & 1.13 & 2.18 & 100 \\
Plural & Affirmative & Without & Name & Present & Irregular & 0.92 & 1.00 & 0.92 & 0.96 & 0.89 & 0.58 & 100 \\
Plural & Affirmative & Without & Pronoun & Past & Regular & 0.63 & 1.00 & 0.63 & 0.77 & 0.10 & 1.92 & 100 \\
Plural & Affirmative & Without & Pronoun & Past & Irregular & 0.68 & 1.00 & 0.68 & 0.81 & 0.57 & 1.64 & 100 \\
Plural & Affirmative & Without & Pronoun & Present & Regular & 0.65 & 1.00 & 0.65 & 0.79 & 1.20 & 2.90 & 100 \\
Plural & Affirmative & Without & Pronoun & Present & Irregular & 1.00 & 1.00 & 1.00 & 1.00 & 1.54 & 0.45 & 100 \\
Plural & Affirmative & With & Name & Past & Regular & 0.88 & 1.00 & 0.88 & 0.94 & 2.94 & 2.44 & 100 \\
Plural & Affirmative & With & Name & Past & Irregular & 1.00 & 1.00 & 1.00 & 1.00 & 3.23 & 2.18 & 100 \\
Plural & Affirmative & With & Name & Present & Regular & 0.60 & 1.00 & 0.60 & 0.75 & 1.00 & 2.48 & 100 \\
Plural & Affirmative & With & Name & Present & Irregular & 0.98 & 1.00 & 0.98 & 0.99 & 1.37 & 0.66 & 100 \\
Plural & Affirmative & With & Pronoun & Past & Regular & 0.91 & 1.00 & 0.91 & 0.95 & 3.02 & 2.20 & 100 \\
Plural & Affirmative & With & Pronoun & Past & Irregular & 0.82 & 1.00 & 0.82 & 0.90 & 3.26 & 2.40 & 100 \\
Plural & Affirmative & With & Pronoun & Present & Regular & 0.67 & 1.00 & 0.67 & 0.80 & 1.42 & 2.58 & 100 \\
Plural & Affirmative & With & Pronoun & Present & Irregular & 1.00 & 1.00 & 1.00 & 1.00 & 2.49 & 0.70 & 100 \\
Plural & Negated & Without & Name & Past & Regular & 0.82 & 1.00 & 0.82 & 0.90 & 1.95 & 1.84 & 100 \\
Plural & Negated & Without & Name & Past & Irregular & 0.91 & 1.00 & 0.91 & 0.95 & 1.79 & 1.26 & 100 \\
Plural & Negated & Without & Name & Present & Regular & 0.91 & 1.00 & 0.91 & 0.95 & 3.09 & 1.94 & 100 \\
Plural & Negated & Without & Name & Present & Irregular & 1.00 & 1.00 & 1.00 & 1.00 & 4.02 & 0.32 & 100 \\
Plural & Negated & Without & Pronoun & Past & Regular & 0.92 & 1.00 & 0.92 & 0.96 & 3.06 & 2.16 & 100 \\
Plural & Negated & Without & Pronoun & Past & Irregular & 1.00 & 1.00 & 1.00 & 1.00 & 2.31 & 1.04 & 100 \\
Plural & Negated & Without & Pronoun & Present & Regular & 0.89 & 1.00 & 0.89 & 0.94 & 3.12 & 2.48 & 100 \\
Plural & Negated & Without & Pronoun & Present & Irregular & 1.00 & 1.00 & 1.00 & 1.00 & 4.42 & 0.24 & 100 \\
Plural & Negated & With & Name & Past & Regular & 0.90 & 1.00 & 0.90 & 0.95 & 2.61 & 1.91 & 100 \\
Plural & Negated & With & Name & Past & Irregular & 0.91 & 1.00 & 0.91 & 0.95 & 1.84 & 1.29 & 100 \\
Plural & Negated & With & Name & Present & Regular & 0.93 & 1.00 & 0.93 & 0.96 & 3.47 & 2.26 & 100 \\
Plural & Negated & With & Name & Present & Irregular & 1.00 & 1.00 & 1.00 & 1.00 & 4.21 & 0.31 & 100 \\
Plural & Negated & With & Pronoun & Past & Regular & 0.85 & 1.00 & 0.85 & 0.92 & 2.54 & 2.22 & 100 \\
Plural & Negated & With & Pronoun & Past & Irregular & 1.00 & 1.00 & 1.00 & 1.00 & 1.92 & 1.29 & 100 \\
Plural & Negated & With & Pronoun & Present & Regular & 0.96 & 1.00 & 0.96 & 0.98 & 3.59 & 1.90 & 100 \\
Plural & Negated & With & Pronoun & Present & Irregular & 1.00 & 1.00 & 1.00 & 1.00 & 4.14 & 0.36 & 100 \\
\bottomrule
\label{tab:appendix_full}
\end{tabular}
}
\caption{Accuracy results on all permutations of settings.}
\label{tab:appendix_full_shrunk}
\end{table}

\begin{table}[h]
\centering
\small
\begin{tabular}{lp{10cm}}
\toprule
\textbf{Setting} & \textbf{Attention Heads in Circuit} \\
\midrule
Base & (11,6), (0,4), (11,4), (0,8), (11,7), (2,6), (1,0), (2,1), (1,1), (6,0), (10,0), (9,4) \\
\midrule
Plural & (11, 6), (0, 4), (11, 4), (0, 8), (11, 7), (2, 6), (1, 0), (2, 1), (1, 1), (6, 0), (10, 0), (9, 4), (11, 11), (11, 1), (10, 7), (0, 6), (0, 10), (1, 8), (9, 6), (11, 3), (4, 4), (6, 5), (11, 5)
\\
\midrule
Negation & (11, 6), (0, 4), (11, 4), (0, 8), (11, 7), (2, 6), (1, 0), (2, 1), (1, 1), (6, 0), (10, 0), (9, 4), (11, 1), (0, 10), (11, 11), (9, 6), (6, 8), (10, 7), (0, 6), (1, 8), (5, 11), (1, 9), (9, 5), (11, 3), (4, 4), (6, 5), (11, 5), (7, 8), (4, 9)
 \\
\midrule
Prefix & (11, 6), (0, 4), (11, 4), (0, 8), (11, 7), (2, 6), (1, 0), (2, 1), (1, 1), (6, 0), (10, 0), (9, 4), (11, 11), (11, 1), (11, 5), (10, 7), (1, 8), (9, 6), (0, 10), (0, 6), (11, 3), (4, 4), (6, 5), (1, 9), (4, 9), (6, 11), (9, 5), (7, 8), (10, 2), (2, 4), (1, 3), (10, 5), (5, 10), (0, 7), (0, 2), (6, 8), (8, 0), (1, 10), (8, 8), (7, 9), (5, 11), (2, 10), (11, 10), (7, 2), (7, 6), (8, 4), (7, 7), (7, 3), (9, 11), (4, 10), (0, 9), (10, 6), (1, 11), (2, 11), (5, 5), (7, 5), (9, 1), (5, 9), (5, 0), (3, 7), (4, 0), (5, 3), (7, 11), (7, 0), (8, 10), (10, 11), (0, 3), (1, 5), (4, 3), (3, 2), (8, 2), (9, 10), (3, 6), (5, 4), (6, 3), (6, 4), (10, 9), (9, 8), (8, 1), (6, 7), (8, 3), (11, 2)
 \\
\midrule
Pronoun & (11, 6), (0, 4), (11, 4), (0, 8), (11, 7), (2, 6), (1, 0), (2, 1), (1, 1), (6, 0), (10, 0), (9, 4), (11, 1), (11, 11), (10, 7), (0, 10), (7, 5), (9, 6), (11, 3), (1, 8), (0, 6), (4, 4)
 \\
\midrule
Past Tense & (11, 6), (0, 4), (11, 4), (0, 8), (11, 7), (2, 6), (1, 0), (2, 1), (1, 1), (6, 0), (10, 0), (9, 4), (11, 11), (10, 7), (0, 10), (11, 1), (1, 8), (8, 8), (0, 6), (11, 3), (6, 5), (4, 4), (4, 9), (11, 5), (7, 8), (9, 6), (1, 9), (6, 11), (5, 10), (10, 5), (10, 2), (0, 2), (9, 5), (1, 3), (6, 8), (0, 7), (2, 4), (5, 11), (7, 9), (1, 10), (5, 5), (2, 10), (9, 11), (11, 10), (7, 2), (7, 6), (8, 0), (8, 4), (4, 10), (3, 7), (7, 11), (7, 7), (2, 11), (0, 9), (7, 5)
\\
\midrule
Irregular & (11, 6), (0, 4), (11, 4), (0, 8), (11, 7), (2, 6), (1, 0), (2, 1), (1, 1), (6, 0), (10, 0), (9, 4), (11, 1), (11, 11), (1, 8), (0, 10), (9, 6), (10, 7), (0, 6), (11, 3), (4, 4), (6, 5), (11, 5), (7, 8), (6, 11), (1, 9), (9, 5), (4, 9), (1, 3), (10, 5), (0, 2), (10, 2), (5, 10), (7, 9), (7, 3), (8, 0), (0, 9), (6, 8), (0, 7), (2, 4), (5, 11), (8, 8), (1, 10), (1, 11), (9, 1), (2, 10), (7, 6), (10, 6), (11, 10), (8, 4), (7, 2), (9, 11), (5, 5), (7, 5), (7, 11), (4, 10), (7, 7), (5, 9), (3, 7), (5, 3), (0, 3), (1, 5), (2, 11), (8, 2), (5, 0), (4, 0), (7, 0), (6, 3), (4, 3), (8, 10), (9, 10), (6, 4), (9, 8), (10, 9), (8, 1), (7, 1), (8, 3), (10, 8), (11, 2), (3, 2), (6, 7), (8, 9), (0, 5), (11, 9), (6, 9), (2, 8), (9, 2), (0, 0), (1, 7), (2, 7), (2, 5), (10, 11), (4, 5), (7, 4), (5, 4), (5, 1), (2, 3)
\\
\midrule
Complex & (11, 6), (0, 4), (11, 4), (0, 8), (11, 7), (2, 6), (1, 0), (2, 1), (1, 1), (6, 0), (10, 0), (9, 4), (11, 11), (11, 1), (0, 10), (10, 7), (1, 8), (0, 6), (4, 4), (9, 6), (11, 5), (6, 11), (6, 5), (11, 3), (4, 9), (1, 9), (0, 2), (9, 5), (10, 5), (8, 10), (7, 8), (1, 3), (10, 2), (5, 10), (8, 8), (2, 4), (6, 8), (5, 11), (7, 9), (1, 10), (0, 7), (2, 10), (11, 10), (8, 4), (9, 10), (7, 2), (7, 7), (9, 11), (10, 8), (7, 6), (8, 0), (0, 9), (3, 10), (7, 11), (4, 10), (1, 11), (5, 5), (7, 5), (5, 9), (2, 11), (9, 1), (7, 3), (3, 7), (8, 2), (5, 0), (4, 0), (5, 3), (10, 6), (2, 8), (7, 0), (2, 5), (0, 3), (6, 3), (1, 5), (10, 9), (6, 4), (9, 8), (8, 1), (6, 7), (11, 2), (3, 2), (0, 5), (8, 3), (6, 9), (11, 9), (8, 9), (0, 0), (5, 4), (2, 7), (9, 2), (10, 11), (4, 5), (7, 4), (5, 1), (4, 3), (8, 6), (8, 7), (1, 7), (8, 11), (3, 6), (7, 10), (10, 3), (6, 2), (9, 9), (1, 2), (5, 6), (9, 3), (3, 1), (3, 4), (10, 4), (5, 7), (4, 11), (11, 0), (4, 7), (4, 2), (3, 3), (9, 7), (6, 6), (1, 6), (3, 0), (6, 1), (4, 6), (10, 10), (8, 5), (3, 8)\\
\bottomrule
\end{tabular}
\caption{Full Circuits Identified per Linguistic Setting}
\label{tab:appendix_full_circuits}
\end{table}

\end{document}